\documentclass[letterpaper]{article} % DO NOT CHANGE THIS
\usepackage{aaai25}  % DO NOT CHANGE THIS
\usepackage{times}  % DO NOT CHANGE THIS
\usepackage{helvet}  % DO NOT CHANGE THIS
\usepackage{courier}  % DO NOT CHANGE THIS
\usepackage[hyphens]{url}  % DO NOT CHANGE THIS
\usepackage{graphicx} % DO NOT CHANGE THIS
\usepackage{natbib}  % DO NOT CHANGE THIS AND DO NOT ADD ANY OPTIONS TO IT
\usepackage{caption} % DO NOT CHANGE THIS AND DO NOT ADD ANY OPTIONS TO IT
\usepackage{algorithm}
\usepackage{algorithmic}
\usepackage{newfloat}
\usepackage{listings}
\DeclareCaptionStyle{ruled}{labelfont=normalfont,labelsep=colon,strut=off} % DO NOT CHANGE THIS
\floatstyle{ruled}
\newfloat{listing}{tb}{lst}{}
\floatname{listing}{Listing}
\usepackage{amsmath}
\usepackage{graphicx}

\title{Targeted View-Invariant Adversarial Perturbations for 3D Object Recognition}
\author{%
  Christian Green  \quad Mehmet Ergezer\thanks{Dr. Ergezer holds concurrent appointments as an Associate Professor at Wentworth Institute of Technology and as an Amazon Visiting Academic. This paper describes work performed at Wentworth Institute of Technology and is not associated with Amazon.}  \quad Abdurrahman Zeybey}
  \affiliations{
  School of Computing and Data Science\\
  Wentworth Institute of Technology\\
  Boston, MA  \\
  \texttt{\{greenc10, ergezerm, zeybeya\}@wit.edu} \\
}

\usepackage{amssymb}
\begin{document}

\maketitle

\begin{abstract}

Adversarial attacks pose significant challenges in 3D object recognition, especially in scenarios involving multi-view analysis where objects can be observed from varying angles. This paper introduces \textbf{View-Invariant Adversarial Perturbations (VIAP)}, a novel method for crafting robust adversarial examples that remain effective across multiple viewpoints. Unlike traditional methods, VIAP enables targeted attacks capable of manipulating recognition systems to classify objects as specific, pre-determined labels, all while using a single universal perturbation. Leveraging a dataset of 1,210 images across 121 diverse rendered 3D objects, we demonstrate the effectiveness of VIAP in both targeted and untargeted settings. Our untargeted perturbations successfully generate a singular adversarial noise robust to 3D transformations, while targeted attacks achieve exceptional results, with top-1 accuracies exceeding 95\% across various epsilon values. These findings highlight VIAP’s potential for real-world applications, such as testing the robustness of 3D recognition systems. The proposed method sets a new benchmark for view-invariant adversarial robustness, advancing the field of adversarial machine learning for 3D object recognition. 

\end{abstract}

\section{Introduction} \label{sec:intro}
Adversarial attacks have emerged as a critical area of research within artificial intelligence, exposing vulnerabilities in machine learning models that can be exploited to manipulate their outputs \cite{carlini2017evaluating}. These attacks, often imperceptible to human observers, pose significant risks to real-world applications, particularly in the context of security \cite{Sharif_2019}. As AI models continue to advance, so too does the urgency of understanding and addressing their susceptibility to adversarial threats \cite{ren2020adversarial}.

Conventional adversarial attacks primarily focus on generating perturbations for single views of objects, exploiting limitations in 2D image understanding \cite{szegedy2013intriguing, madry2017towards}. These attacks face limitations when applied to 3D objects, as they often lack robustness to real-world variations in perspective. Unlike 2D systems, 3D recognition systems must contend with viewpoint variability, transformations, and real-world conditions, making it difficult to craft perturbations that generalize effectively across multiple perspectives.

This paper introduces \textbf{View-Invariant Adversarial Perturbations (VIAP)}, a novel framework designed to overcome these challenges by generating robust, universal perturbations for 3D objects. VIAP enables adversarial attacks that maintain their effectiveness across diverse transformations of a 3D object \cite{zhi2018toward}, including rotations and viewpoint changes, while also allowing targeted manipulation of class labels \cite{athalye2018synthesizing}. Unlike prior methods that primarily focus on non-targeted attacks \cite{ergezer2024noise}, VIAP extends adversarial capabilities to precise, controlled targeted attacks, broadening its practical applicability.

\subsubsection{Problem and Contributions}
Adversarial attacks in 3D object recognition confront a fundamental challenge: developing perturbations that remain effective across diverse viewpoints. Existing adversarial perturbation methods suffer from two critical limitations: (1) poor generalizability across multi-angle views, and (2) restricted capability for targeted attacks. Our work directly addresses these constraints through three key contributions:

\begin{itemize}
    \item A novel method for generating view-invariant universal perturbations that maintain effectiveness across multiple object perspectives.
    \item A mathematical formalization of view-invariance in adversarial attacks, providing theoretical grounding for multi-view robustness.
    \item Experimental validation demonstrating superior performance in both targeted and untargeted scenarios, outperforming established baseline methods.
\end{itemize}

By introducing VIAP, we bridge critical gaps in current adversarial machine learning approaches, enabling more sophisticated and reliable attacks that can adapt to complex 3D recognition environments.

% \subsubsection{Contributions:}
% \begin{itemize}
%     \item We propose a novel method for generating targeted universal perturbations for 3D objects that remain effective across multiple viewpoints.
%     \item We mathematically formalize the concept of view-invariance in adversarial attacks.
%     \item Our method demonstrates superior performance in both targeted and untargeted scenarios compared to established baseline methods.
% \end{itemize}

% \subsubsection{Problem Statement:}
% Adversarial attacks in 3D object recognition face challenges due to viewpoint variability. Existing universal perturbation methods are limited in their ability to generalize across multi-angle views, particularly for targeted attacks. This work addresses this limitation by proposing a targeted view-invariant perturbation method that adapts to these challenges.

In the following sections, we provide a review of related work to contextualize our contributions and identify existing gaps in adversarial research. We then present our targeted View-Invariant Adversarial Perturbation (VIAP) method, detailing the mathematical foundation and implementation. Extensive experiments validate the robustness and versatility of our approach in both targeted and untargeted scenarios, demonstrating its superiority over baseline methods. Finally, we discuss practical implications, limitations, and directions for future work, emphasizing the broader impact of this study on the security of 3D object recognition systems.

\section{Related Work} 

This section reviews foundational adversarial attack methods, highlighting their principles and limitations, which set the stage for the proposed targeted View-Invariant Adversarial Perturbations approach discussed in the next section. %~\ref{sec:methods}.

\subsubsection{Fast Gradient Sign Method (FGSM):} One of the earliest adversarial attack methods, was proposed in 2014 \cite{goodfellow2015explaining}. FGSM quickly gained popularity due to its simplicity, low computational cost, and ease of implementation. The method relies on attacks where the assailant knows the model’s architecture and weights. However, this reliance limits FGSM’s effectiveness in settings where only the input features are accessible.

FGSM works by calculating the gradient of the loss with respect to the input image and then updating the image's pixels in a direction that maximizes this loss. In  Equation~\ref{eq:fgsm}, we let $X$ and $y$ be the input image and true label respectively. We assume $X$ as a 3-D matrix (width × height × color). The perturbation strength is controlled by \(\epsilon\). \(\nabla_X J(X, y_{true})\) is the gradient of the model's loss with respect to \(X\). The sign of the gradient determines the direction of the perturbation. We initialize $X^{adv}_0 = X$.

\begin{equation}\label{eq:fgsm}
X^{adv} = X + \epsilon  \cdot\text{sign}( \nabla_X J(X, y_{true}))   
\end{equation}

While FGSM is primarily used for untargeted attacks, it can be adapted for targeted attacks by adjusting the gradient direction to minimize the loss between the targeted and predicted labels, as shown in Equation~\ref{eq:fgsmtar}.

\begin{equation}\label{eq:fgsmtar}
X^{adv} = X + \epsilon \cdot \text{sign}( \nabla_X J(X, y_{target}))  
\end{equation}

\subsubsection{The Basic Iterative Method (BIM)} Extends the idea of FGSM by applying the perturbation iteratively, allowing for a more effective adversarial attack \cite{kurakin2017adversarial}. By clipping intermediate results at each step, BIM ensures that the perturbation stays within a defined range, as outlined in Equation~\ref{eq:bim}. This iterative approach typically enhances the success rate of the attack compared to the single-step FGSM. As with FGSM, we initialize BIM, $X^{adv}_0 = X$. $N$ in Equation~\ref{eq:bim} represents the iteration count. 

\begin{equation}\label{eq:bim}
X_{N+1}^{adv} = \text{Clip}_{X,\epsilon} \{X_{N}^{adv} + \epsilon  \cdot\text{sign}( \nabla_X J(X_{N}^{adv}, y_{true}))\}
\end{equation}

\subsubsection{Untargeted View-Invariant Adversarial Perturbations:} The conventional process of generating an adversarial noise involves crafting a unique perturbation for an image. The noise is applied to the image and fed through a classifier, leading to an incorrect class label prediction. Attempting to apply this perturbation to separate images often leads to inconsistent outputs by our classifier, rendering the noise useless.

The previous iteration of VIAP, published as the universal perturbation method, was created to overcome this obstacle \cite{ergezer2024noise} by formulating one perturbation universally applicable to various perspectives of one or more objects, $\mathbb{X}^{\text{adv}}$. To achieve this, a modification was made to the BIM. The gradients are calculated with respect to the adversarial noise itself, instead of the input image. This adjustment separated the number of input images from the shape of the generated adversarial noise. In turn, it allowed the attacker to control the input image's dimensions and the associated noise, independently. This update enables simultaneous inputs of various views for the same object. The new algorithm provided solitary adversarial noise capable of collectively compromising the recognition of all these distinct perspectives at once (including rotations, changes in lighting, and potential deformations).

Equation~\ref{eq:Universal} details VIAP calculation where $X$ is stacked input images as a 4-D tensor (image count, width, height, color), $\mathbb{X}$ is the calculated perturbation as a 3-D matrix (width, height, color), matching the shape of an individual image in $X$.  $N$ is the number of desired iterations as defined in BIM. $\epsilon$ is a hyperparameter controlling the scale of the attack as used in FGSM. \( \nabla_{\mathbb{X}_{N}} J(\mathbb{X}_{N}^{\text{adv}}, y_{\text{true}}) \) is the gradient of the model calculated using a cross-entropy loss function with respect to the generated perturbation.

\begin{equation}
\begin{aligned}
\mathbb{X}_{N+1}^{\text{adv}} &= \text{Clip}_{X,\epsilon} \{\mathbb{X}_{N}^{\text{adv}} + \epsilon \cdot \text{sign}(\nabla_{\mathbb{X}_{N}} J(\mathbb{X}_{N}^{\text{adv}}, y_{\text{true}}))\}
\end{aligned} \label{eq:Universal}
\end{equation}

Unlike FGSM and BIM, the perturbation is initialized as $\mathbb{X}^{adv}_0 = X + r$ where $r\sim U(-0.01, 0.01)$.

\subsubsection{Multi-Object vs Single-Object Universal Perturbations:} Prior work on universal perturbations has primarily focused on generating adversarial noise that can fool classifiers across different objects and classes. \cite{universal_2017_CVPR} introduced universal perturbations that could mislead classification across diverse images from different categories. Their method aims to find a single perturbation vector that can cause misclassification when applied to any image from the training distribution. Similarly, \cite{poursaeed2018generative} extended this concept to generative universal adversarial perturbations.

% Our approach fundamentally differs from these previous works in both objective and methodology. While traditional universal perturbations seek to find a single noise pattern that works across different objects, our method focuses on generating perturbations that remain effective across multiple views of the same 3D object. This distinction is crucial for several reasons:

% 1. View Consistency: Our method explicitly optimizes for perturbation consistency across different perspectives of the same object, ensuring the attack remains effective as the viewing angle changes. Traditional universal perturbations do not consider this view-invariance constraint.

% 2. Geometric Preservation: By focusing on a single object, our method can better preserve the geometric relationships between different viewpoints, maintaining attack effectiveness while respecting the underlying 3D structure. This is particularly important for real-world applications where objects are viewed from multiple angles.

% 3. Target Specificity: Our approach allows for more precise control over the targeted attack outcome while maintaining effectiveness across viewpoints. Previous universal perturbation methods typically focus on untargeted attacks or require different perturbations for different target classes.

Consider a security camera system monitoring a 3D object from multiple angles. Traditional universal perturbations might require different adversarial patterns for each view, potentially breaking the attack's effectiveness as the camera angle changes. Our method addresses this limitation by generating a single perturbation that remains robust across viewpoints, making it more practical for real-world adversarial attacks on 3D object recognition systems.

This geometric consideration also enables our method to achieve higher success rates with smaller perturbation magnitudes compared to traditional universal perturbations when applied to multi-view scenarios. As demonstrated in our experimental results, this approach achieves superior performance in both targeted and untargeted settings while maintaining consistency across different viewpoints of the same object.

\section{Methodology \label{sec:methods}}

In this section, we introduce the targeted View-Invariant Adversarial Perturbations method and detail the enhancements made to the existing algorithm which allow it to perform in targeted environments. We first describe the dataset and preprocessing steps, followed by the specifics of the perturbation generation process, and conclude with the evaluation metrics and experimental setup.

\subsection{Dataset and Preprocessing}
%CHRIS: TODO Add reference
Our experiments utilize a dataset comprising 1,210 images of 121 distinct 3D objects, each rendered from multiple viewpoints. The objects were selected to cover a diverse range of categories, ensuring the robustness of the proposed method across different types of 3D shapes and textures. 3D models were downloaded for six objects
\cite{baseball_cite},
\cite{snail_cite},
\cite{acorn_cite},
\cite{conch_cite},
\cite{pretzel_cite},
\cite{lemon_cite},  
\cite{broccoli_cite}, 
\cite{tractor_cite},
and the rest were obtained through a dataset \cite{objaverse}. Each of the 121 objects fell under one of 14 distinct class labels:
baseball,
snail,
acorn,
conch,
pretzel,
lemon,  
broccoli, 
tractor,
backpack,
banana,
dumbbell,
pineapple,
strawberry,
and teddy bear. 

\begin{figure}[]
    \centering
    \includegraphics[width=3.6in]{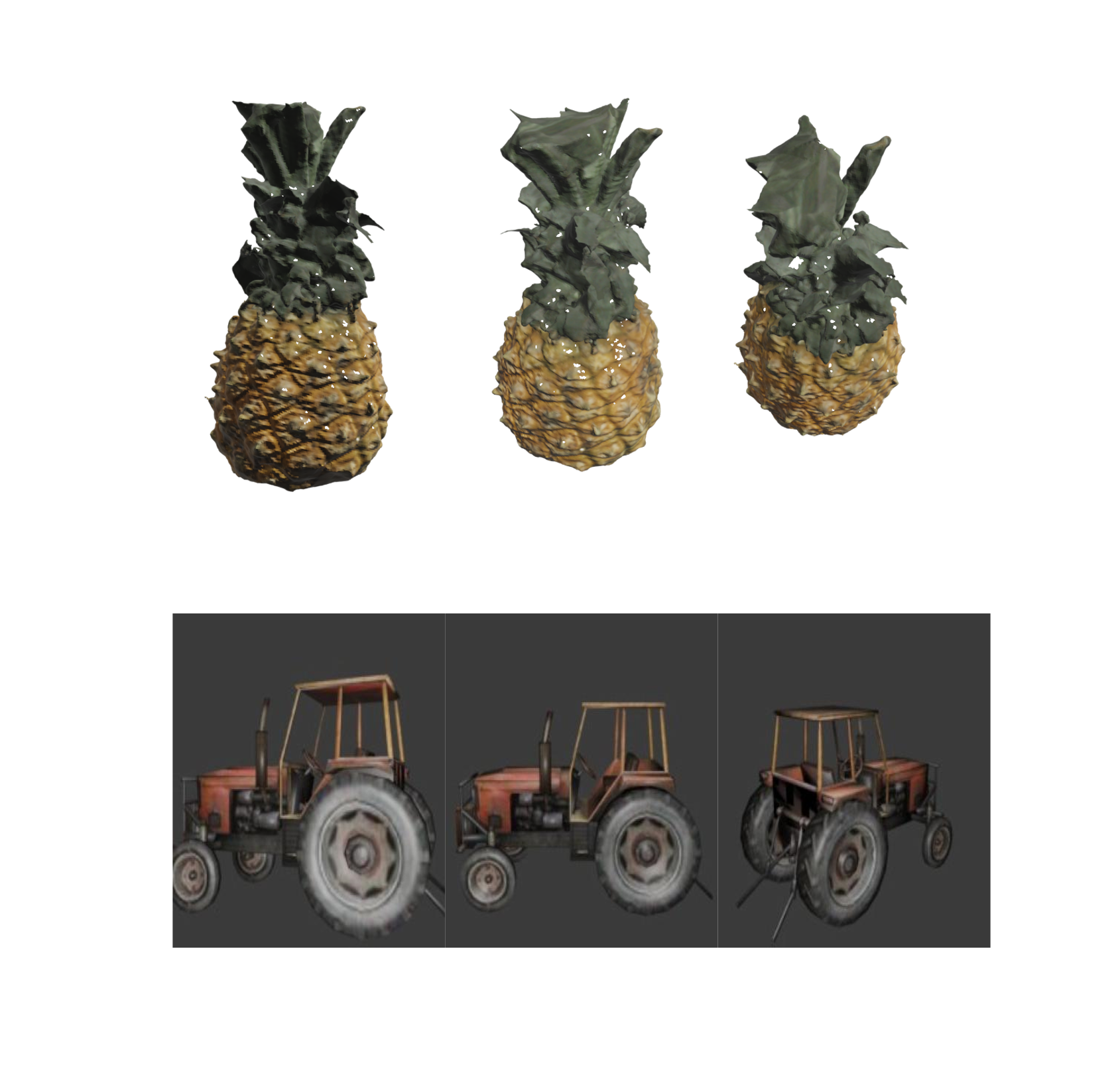}
    \caption{The top row consists of three rendered camera angles of a pineapple from the 3D object dataset. The bottom row consists of three rendered images of a tractor from Blender.}
    \label{img:rendered}
\end{figure}

\medskip \textbf{Image Rendering:}
Each object was rendered from 10 different viewpoints, simulating real-world conditions where objects may be viewed from various angles. Renders of random camera orientation for two of the objects are shown in Figure~\ref{img:rendered}.

\medskip \textbf{Preprocessing:} The images were resized to a standard resolution, $(224,224)$ pixels, to maintain consistency across the dataset. Standard preprocessing techniques such as normalization were applied to the images to ensure they are suitable inputs for the neural networks. 

\subsection{Mathematical Formalization of View-Invariance}
To quantify the view-invariance property of the proposed targeted perturbation, we define a set of transformations \( \mathcal{T} \) that represent the changes in viewpoint, including rotation, translation, and scaling. Let \( X_{i}^{(\theta, \phi)} \) denote an image of object \( X_{i} \) rendered at a viewpoint specified by angles \( (\theta, \phi) \). The targeted VIAP \( \delta \) is optimized to ensure that for any transformation \( T \in \mathcal{T} \), the following holds:
\[
\arg\max_{c} f\left( X_{i}^{(\theta, \phi)} + \delta \right) = y_{\text{target}}, \quad \forall (\theta, \phi) \in \Theta \times \Phi,
\]
where \( f(\cdot) \) represents the classifier function and \( y_{\text{target}} \) is the target class label.

To achieve this, we modify the gradient calculation to incorporate the distribution over multiple transformations:
\[
\nabla_{\delta} J(\delta) = \mathbb{E}_{T \sim \mathcal{T}} \left[ \nabla_{\delta} J(f(X_{i}^{(T)} + \delta), y_{\text{target}}) \right].
\]

This ensures that the perturbation \( \delta \) remains effective under different viewpoints, capturing the multi-view robustness of the attack.

\subsection{Targeted View-Invariant Adversarial Perturbations}

We extend the View-Invariant Adversarial Perturbations method, from Equation~\ref{eq:Universal} to incorporate targeted attacks, allowing for precise adversarial manipulation. Targeted attacks are achieved by computing the cross-entropy loss between the target label and the predicted adversarial example. The gradient of the loss with respect to the calculated perturbation was then multiplied by a negative one. This ensures that through each training iteration, the perturbation loss is minimized. This is formulated in Equation~\ref{eq:Universaltarget} and in Algorithm~\ref{alg:targettedUP} where Clip limits the perturbed image values. 

\begin{equation}
\begin{aligned}
\mathbb{X}_{N+1}^{\text{adv}} &= \text{Clip}_{X,\epsilon} \{\mathbb{X}_{N}^{\text{adv}} - \epsilon \cdot \text{sign}(\nabla_{\mathbb{X}_{N}} J(\mathbb{X}_{N}^{\text{adv}}, y_{\text{target}}))\}
\end{aligned} \label{eq:Universaltarget}
\end{equation}

\begin{algorithm}
\caption{Targeted View-Invariant Perturbation} \label{alg:targettedUP}
\begin{algorithmic}
% \SetAlgoHangIndent{0pt}

\STATE \texttt{INPUT} Dataset \( X \), target label \( y_{\text{target}} \), step size \( \epsilon \), max iterations \( N \)
\STATE \texttt{OUTPUT} VIAP \( \delta \) 
\STATE Initialize \( \delta \) with small random noise \( \delta_{0} \sim U(-0.01, 0.01) \)
\FOR {each iteration \( n \) from 1 to \( N \)}
\STATE Compute gradient: \( g = \nabla_{\delta} J(f(X + \delta), y_{\text{target}}) \) 
\STATE Update perturbation: \( \delta = \text{Clip}_{\epsilon}(\delta - \eta \cdot \text{sign}(g)) \)
\ENDFOR
\STATE \RETURN \( \delta \)
\end{algorithmic}
\end{algorithm}

% \begin{algorithm}[H]
% % \SetAlgoHangIndent{0pt}
% \caption{Targeted View-Invariant Perturbation:}
% \KwIn{Dataset \( X \), target label \( y_{\text{target}} \), step size \( \epsilon \), max iterations \( N \)}
% \KwOut{VIAP \( \delta \)}
% Initialize \( \delta \) with small random noise \( \delta_{0} \sim U(-0.01, 0.01) \)
% \For{each iteration \( n \) from 1 to \( N \)}{
% Compute gradient: \( g = \nabla_{\delta} J(f(X + \delta), y_{\text{target}}) \) \\
% Update perturbation: \( \delta = \text{Clip}_{\epsilon}(\delta - \eta \cdot \text{sign}(g)) \)
% }
% Return \( \delta \)
% \label{alg:targettedUP}
% \end{algorithm}

\subsection{Experimental Setup}

To showcase the effectiveness of our model in generating robust noise across multiple angles, we conducted experiments comparing our View-Invariant Adversarial Perturbations method against FGSM and BIM.

% Tommy - hide the original paragraphs written below this section (hidden within \iffalse). You can replace the paragraph I reorganized with the original one if you feel like the original one looked better.
We start by rendering our 3D objects in Blender to gain multiple 2D images from different viewpoints. Depending on the object, we take 2D images straight from a 3D rendered dataset. Eight of the objects were rendered from ten distinct viewing angles, ensuring consistent recognition by our classification model across perspectives. The remaining objects were gathered from 3D render datasets consisting of multiple images of an object at different angles. For the rendered images gathered through Blender, camera angles were randomly selected based on a spherical coordinate system centered around the object, incorporating a 15\% random deviation for robustness. The slight changes in the coordinates mimicked natural variations in viewing angles encountered in real-world environments. Consequently, constant lighting positions relative to the camera angles generated shadowed areas that hindered initial MobileNetV2 recognition. To combat such challenges it was necessary to restrict the 15\% deviation on certain objects to one axis.

The images are split into two sets, one training and one test. A single perturbation is developed, applicable to all images in the training set. The noise is added to all images before inferring with MobileNetV2 \cite{sandler2019mobilenetv2}. The test set is used to evaluate the generalizability of the perturbation. All steps are repeated for the BIM and FGSM attacks. Each experiment is done twice to incorporate both targeted and untargeted use cases. Before running any experiments, we gathered baseline results to ensure the images were properly classified without any adversarial noise.

In cases of targeted attacks, we reduced biases by randomly selecting a label from the $1,000$ classes our classifier was trained on for each trial of our experiments. If the randomized target label was chosen to be equal to the true label, we randomly sampled another label.

\subsection{Evaluation Metrics} 
%CHRIS: TODO

To assess the effectiveness of the targeted View-Invariant Adversarial Perturbations, we employ several key metrics:

\medskip \textbf{Top-1 Accuracy}: Equation~\ref{eq:top1acc} evaluates the attack success by measuring the drop in top-1 accuracy for both untargeted and targeted perturbations. For targeted attacks, Equation~\ref{eq:top1acctar} monitors the accuracy of the model in misclassifying images as the targeted class.
\setlength{\belowdisplayskip}{0pt} \setlength{\belowdisplayshortskip}{0pt}
\setlength{\abovedisplayskip}{0pt} \setlength{\abovedisplayshortskip}{0pt}

\begin{equation} \label{eq:top1acc}
\text{Top-1 Acc} = \frac{\Sigma \text{ Top-1 True Label Predicts}}{\Sigma \text{ Total Predicts}}
\end{equation}

\begin{equation} \label{eq:top1acctar}
\text{Top-1 Target Acc} = \frac{\Sigma \text{ Top-1 Target Label Predicts}}{\Sigma \text{ Total Predicts}}
\end{equation}

% \begin{equation} \label{eq:top1acc}
% \begin{aligned}
% \text{Top-1 Accuracy} = \frac{ \text{\# of Top-1 True Label Classifications} }{ \text{Total \# of Classifications}}
%  \end{aligned} 
%  \end{equation}

%  \begin{equation}
%  \begin{aligned}
% \text{Top-1 Target Accuracy} = \frac{ \text{\# of Top-1 Target Label Classifications} }{ \text{Total \# of Classifications}}
% \end{aligned} \label{eq:top1acctar}
% \end{equation}

\medskip \textbf{Perturbation Robustness:} We test the perturbation’s robustness by applying it to unseen viewpoints and verifying its effectiveness.

\begin{figure}[]
    \centering
    \includegraphics[width=3.0in]{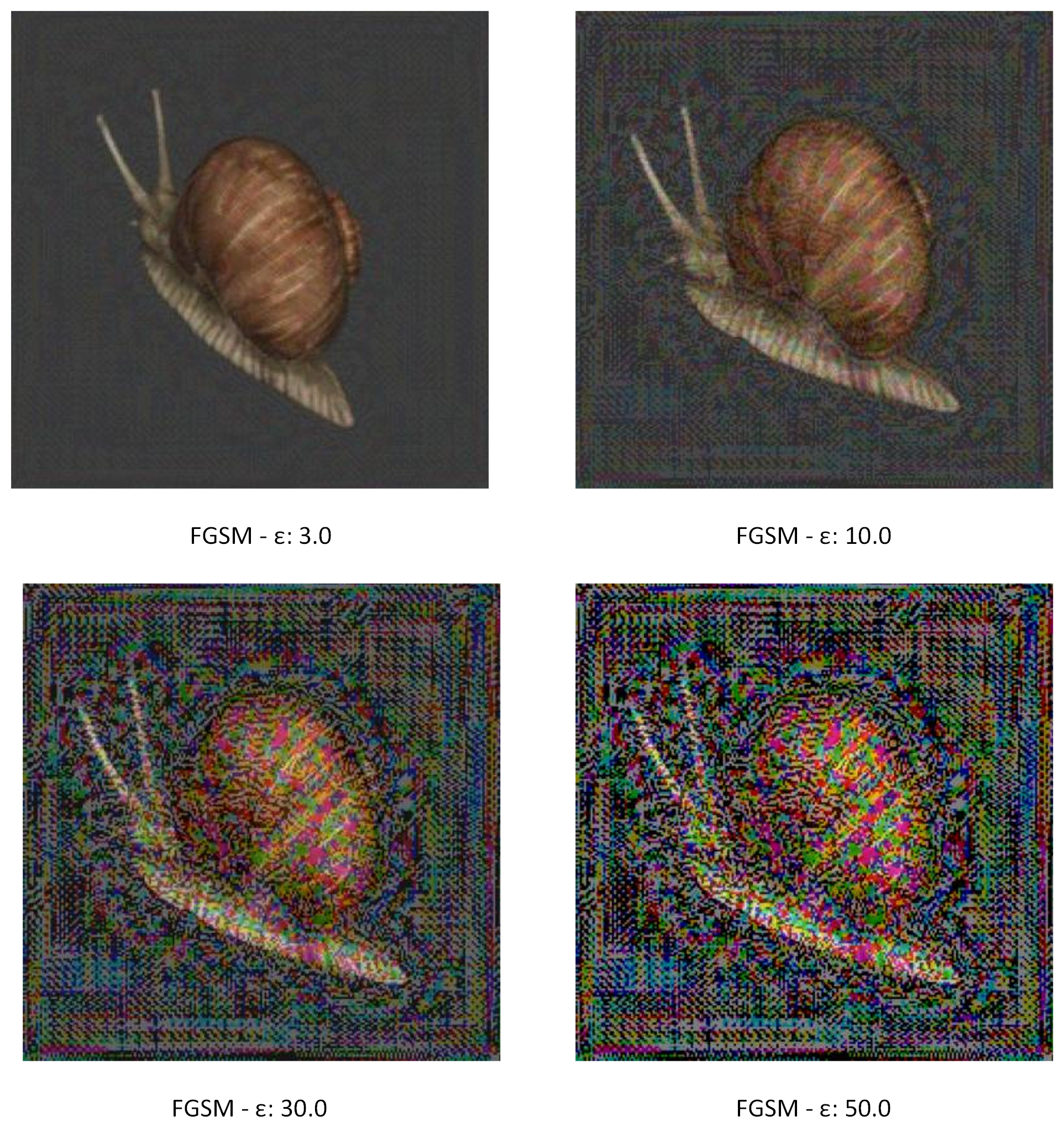}
    \caption{FGSM attacks at increasing $\epsilon$ values.}
    \label{img:snail_eps_compare}
\end{figure}

\medskip \textbf{Parameter Selection Rationale:} Parameter \( \epsilon \) controls the perturbation magnitude and is chosen based on a balance between attack strength and imperceptibility. Through preliminary experiments, we selected \( \epsilon \) values that demonstrated significant classification drops without perceptual distortion. Figure~\ref{img:snail_eps_compare} illustrates the consequence of increasing $\epsilon$ values on the distortion of the original image. 

\medskip \textbf{$\epsilon$ Sensitivity Analysis:} By varying $\epsilon$, we analyze the trade-off between perturbation strength and attack success, identifying the optimal $\epsilon$ range for both targeted and untargeted attacks. We perform experiments on our data with epsilon values ranging between 0 and 50.

% \subsection{View-Invariant Robustness Metric}
% We introduce a metric \( R_{\text{view}} \) to evaluate the robustness of the perturbation against viewpoint variations:
% \[
% R_{\text{view}}(\delta) = \frac{1}{|\Theta| \times |\Phi|} \sum_{(\theta, \phi) \in \Theta \times \Phi} \mathbb{1}\left[ \arg\max_{c} f\left( X_{i}^{(\theta, \phi)} + \delta \right) = y_{\text{target}} \right],
% \]
% where \( \mathbb{1}[\cdot] \) is the indicator function. A higher \( R_{\text{view}} \) value indicates that the perturbation is successful across a broader range of views.

\section{Results}\label{sec:exp}

%CHRIS: TODO: copy figures from thesis as supplement 4.1/4.2 - 4.5/4.6

%CHRIS: TODO: figure 4.4 in text directly but add the attack result? + non-attacked? 
% Another object? 

\begin{figure*}[]
    \centering
    \includegraphics[width=6.0in]{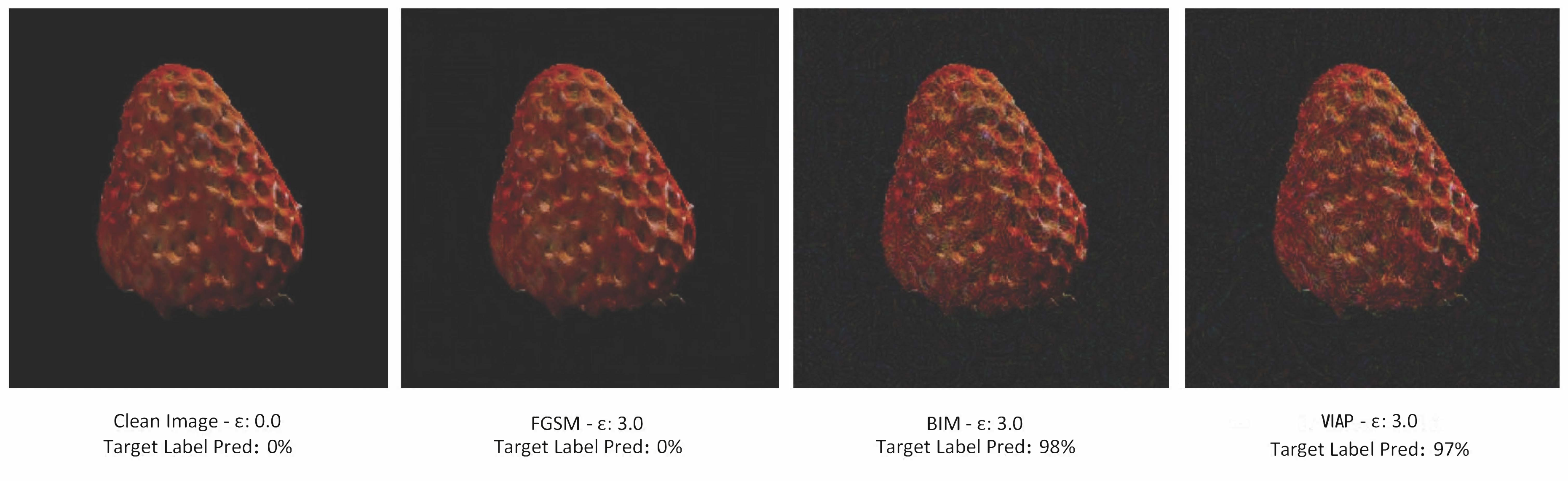}
    \caption{Figure consists of four rendered images of a strawberry with each adversarial attack at the same epsilon value. The target label (oscilloscope) soft-max prediction for each image is also listed. The clean image is included as a basis. The perturbation is still quite transparent at \(\epsilon\) = 3.0. }
    \label{img:strawberry_eps_compare}
\end{figure*}

To showcase the effectiveness of our model in generating robust noise across multiple angles, we conducted experiments comparing our View-Invariant Adversarial Perturbations method with common adversarial attack methods: FGSM and BIM. The objective was to identify the optimal noise level that significantly reduces a classifier's confidence in the object's true class while also increasing the target class' confidence. %By accomplishing these tasks, we aimed to validate our proposed method.

%on a vast number of 3D objects, while also improving the given algorithm to allow control of the attack's output.

\subsection{Untargeted Attack Confidence}
%CHRIS: TODO: drop decimal places

Table~\ref{tab:untarget_table} summarizes the average softmax predictions of the true object label when images are subjected to different adversarial attacks. We analyze the robustness of these attacks across 605 training and 605 unseen images. BIM and our View-Invariant Adversarial Perturbations method both adhere to 20 iterations while FGSM is limited to one iteration. 

\begin{table*}[t]
% \begin{table}[ht]
\centering
\begin{tabular}{|l||ccc||ccc|}
\hline
\multicolumn{1}{|c||}{$\epsilon$} & \multicolumn{3}{c||}{Train Images $\times 10^{-6}$} & \multicolumn{3}{c|}{Test Images $\times 10^{-6}$} \\ \cline{2-7} 
            & \phantom{ }FGSM    & BIM      & VIAP\phantom{ }   & \phantom{ }FGSM    & BIM      & VIAP \phantom{ } \\ \hline
        0.00 & 509,757 & 509,757 & 496,621 & \textbf{519,231} & \textbf{519,231} & \textbf{507,621} \\ 
        0.50 & 78,091 & 0 & 16 & 491,976 & 258,959 & \textbf{78,269} \\ 
        1.00 & 35,697 & 0 & 0 & 472,773 & 143,625 & \textbf{26,230} \\ 
        3.00 & 21,658 & 0 & 0 & 446,958 & 10,037 & \textbf{2,386} \\ 
        5.00 & 20,657 & 0 & 0 & 394,593 & 444 & \textbf{3} \\ 
        10.00 & 17,980 & 0 & 0 & 230,679 & \textbf{2} & 262 \\ 
        15.00 & 12,265 & 0 & 0 & 109,324 & \textbf{38} & 5,403 \\ 
        30.00 & 1,651 & 0 & 0 & 4,924 & \textbf{1} & 49\\ 
        50.00  & 464 & 3 & 1 & 674 & \textbf{0} & 3 \\                           
 \hline
Mean        & 77,580  & 56,640	 & 55,182	     & 296,793	 & 103,590	 & \textbf{68,279} \\
Std $(\pm)$ & 163,718	 & 169,919 & 165,539	 & 213,028	 & 180,558	 & \textbf{166,786} \\
 \hline

\end{tabular}
\caption{Average soft max predictions of the true label from MobileNetV2 under different untargeted adversarial attacks and epsilon values for both train and test images. $\epsilon=0.00$ indicates clean images without adversarial noise. The lower the result, the more successful the attack.}
\label{tab:untarget_table}
\end{table*}

In Table~\ref{tab:untarget_table}, our proposed and basic iterative attacks provide the best success at the lowest noise levels. At $\epsilon= .50$ all three attacks exhibit impressive performance on their training images. Although the difference is marginal, both the BIM and VIAP attacks have the lowest average soft-max predictions. This pattern can be seen through all epsilon values from .5 to 30. At $\epsilon = 30$ each attack sees a slight shift in performance on the training images. BIM and VIAP jump from 0.00 to 0.22E-06 and 0.00 to 0.24E-06, respectively. FGSM also sees improvement in its average softmax prediction, dropping from 12265.05E-06 to 1651.91E-06. Similar changes are reflected at $\epsilon = 50$.

For the test images, the best attack result for each $\epsilon$, corresponding to the lowest softmax prediction of the true label, is highlighted in bold. While the performance on the training set is exceptional, the results for the test images show a significant degradation in the effectiveness of each attack. From $\epsilon = .50$ to $\epsilon = 5.00$, our VIAP method displays the most success among the three attacks. At $\epsilon = 10$ and beyond, BIM proves to provide the most effective perturbation. The decline in adversarial performance between train and test data is expected since the noise is being evaluated on images it hasn't seen before. Despite these changes in results, our VIAP attack is still more than adequate for practical applications. Even at its least effective epsilon value ($\epsilon = .50$),  our proposed attack produces an average softmax prediction of ~8\%, which is low enough to deceive our classifier. The proposed attack's achievements are extended through its robustness to different views.

\subsection{Targeted Attack Confidence}

Table~\ref{tab:target_table} summarizes the average softmax predictions of the \textbf{target label} when images are subjected to different adversarial attacks. Our proposed targeted View-Invariant Adversarial Perturbations method and BIM provide the highest success on both train and test images. At $\epsilon = 0.00$, all attacks have a softmax value of 0 for train and test images since no adversarial noise was added, and MobileNetV2 classifies the image as the true label instead of the target label. For train images at $\epsilon = 0.5$, BIM and targeted View-Invariant Adversarial Perturbations exhibit noteworthy performance with average predictions of 0.91 and 0.73, respectively. The softmax prediction of both attacks gradually increases until reaching a maximum value of 0.97 at $\epsilon = 5.0$. Past that point, both attacks decline until reaching a minimum of 0.12 for BIM and 0.16 for targeted VIAP. FGSM does not successfully match up with the other attacks at any $\epsilon$ value, having a maximum average prediction of .01 between $\epsilon = 1.0$ - $5.0$.

%CHRIS: TODO stat significance test? p-test or similar
% \begin{table}[ht]
\begin{table*}[t]
\centering
\begin{tabular}{|l||ccc||ccc|}
\hline
\multicolumn{1}{|c||}{$\epsilon$} & \multicolumn{3}{c||}{Train Images} & \multicolumn{3}{c|}{Test Images} \\ \cline{2-7} 
            & \phantom{ }FGSM    & BIM      & VIAP\phantom{ }   & \phantom{ }FGSM    & BIM      & VIAP \phantom{ } \\ \hline
        0.00 & 0.00 & 0.00 & 0.00 & \textbf{0.00} & \textbf{0.00} & \textbf{0.00} \\ 
        0.50 & 0.00 & 0.91 & 0.73 & 0.00 & 0.01 & \textbf{0.09} \\ 
        1.00 & 0.01 & 0.97 & 0.94 & 0.00 & 0.04 & \textbf{0.31} \\ 
        3.00 & 0.01 & 0.97 & 0.97 & 0.00 & 0.18 & \textbf{0.61} \\ 
        5.00 & 0.01 & 0.97 & 0.97 & 0.00 & 0.25 & \textbf{0.74} \\ 
        10.00 & 0.00 & 0.91 & 0.94 & 0.00 & 0.29 & \textbf{0.76} \\ 
        15.00 & 0.00 & 0.81 & 0.88 & 0.00 & 0.27 & \textbf{0.71} \\ 
        30.00 & 0.00 & 0.40 & 0.45 & 0.00 & 0.01 & \textbf{0.32}\\ 
        50.00  & 0.00 & 0.12 & 0.16 & 0.00 & 0.03 & \textbf{0.11} \\                           
 \hline
Mean & 0.01  & 0.67	 & 0.67	 & 0.00	 & 0.13	 & \textbf{0.41} \\
Std $(\pm)$ & 0.01	 & 0.39	 & 0.38	 & 0.00	 & 0.12	 & \textbf{0.30} \\
 \hline

\end{tabular}
\caption{Average softmax predictions of MobileNetV2 under different targeted adversarial attacks and epsilon values for both train and test images. $\epsilon=0.00$ indicates clean images without an attack. The higher the result, the more successful the attack.}
\label{tab:target_table}
\end{table*}

Regarding the test images, the FGSM result remains nearly constant at 0.00 for every $\epsilon$ value, meaning FGSM does not produce successful adversarial examples at any noise level. Targeted VIAP and BIM both perform poorly on the test data at $\epsilon = 0.50$, with average predictions of 0.09 and 0.01, respectively. The overall trend of these attacks shows increased success at larger $\epsilon$ values with targeted VIAP performing best with a softmax prediction of 0.76 at $\epsilon = 10.0$. These results confirm that our method produces stronger perturbations than the FGSM attack at each noise level, especially for test data where our method shows a mean softmax prediction nearly four times as high as FGSM. Our attack displays stronger generalization to unseen images compared to the BIM attack as well, proving its potential as a practical application.

\subsection{Statistical Significance of Targeted VIAP Results}

To assess the statistical significance of the targeted adversarial test results, we conducted two-sample t-tests between the softmax predictions generated by our View-Invariant Adversarial Perturbations method, FGSM, and BIM. The p-values for the comparisons are reported in Table \ref{table:pvalues}. With a significance threshold set at $\alpha = 0.05$, the View-Invariant Adversarial Perturbations method yielded a p-value of 0.0005 compared to FGSM and 0.0095 compared to BIM. These results indicate a statistically significant difference in performance between our method and both FGSM and BIM, with a stronger significance observed when compared to FGSM. This reinforces the effectiveness of our View-Invariant Adversarial Perturbations method in producing stronger adversarial examples, particularly in targeted attack scenarios.

\begin{table}[h!]
\centering
\begin{tabular}{c|c}
%\hline
VIAP-FGSM & VIAP-BIM   \\ \hline
0.0005 & 0.0095 \\ %\hline
\end{tabular}
\caption{P-values from the t-tests comparing the proposed VIAP method to FGSM and BIM.}
\label{table:pvalues}
\end{table}

\section{Discussion}
The results of our study have significant implications for adversarial attack and defense strategies in multi-view recognition systems. This work highlights the potential for robust adversarial manipulation in real-world applications by demonstrating the effectiveness of targeted VIAP perturbations across diverse viewpoints. 

% \textbf{Limitations} of our approach include its reliance on synthetic 3D data, which may not fully capture the complexity of real-world scenarios. Future work should focus on extending this method to real-world datasets and exploring adaptive defenses capable of mitigating such robust adversarial perturbations.
\medskip \textbf{Limitations Discussion:}
Our current method, while robust in multi-view synthetic data, may exhibit reduced effectiveness when applied to complex, real-world 3D environments due to unmodeled factors such as texture variations and lighting inconsistencies. Future work will explore adaptive training techniques and real-world dataset validation to address these limitations.

Potential applications of this method include evaluating the robustness of security systems, enhancing adversarial training processes, and benchmarking multi-view recognition algorithms under adversarial conditions. Future studies could also examine the interplay between perturbation strength and visual perceptibility to balance efficacy with detectability.

\section{Conclusion}

This paper introduced a novel approach for generating View-Invariant Adversarial Perturbations and optimizing adversarial robustness across various perspectives. By focusing on 3D-rendered images of objects viewed from multiple angles, we emphasized the method's utility in practical applications. The experimental results validate the effectiveness of our approach in both targeted and untargeted scenarios, with notable success in the targeted category, achieving a top-1 accuracy exceeding 95\% for most tested epsilon values.

Our method demonstrates superior performance in comparison to established techniques like FGSM and BIM, particularly in terms of transferability and generalization to unseen data.  
Furthermore, we successfully extend our proposed algorithm's applicability to targeted use cases, establishing its capability for more controlled adversarial alterations. Results indicate that between \(\epsilon\) = 0.5 and \(\epsilon\) = 5.0, the untargeted VIAP attack displays the highest attack success rate. Targeted VIAP attacks exhibited the best top-1 target accuracy for test images while requiring the least amount of computational effort. The importance of the View-Invariant Adversarial Perturbations method is highlighted by its ability to operate exclusively on 2D images, offering a practical and scalable alternative to computationally costly 3D adversarial attacks.

In future work, we plan to extend our experiments to real-world 3D objects and explore the application of this method to other adversarial scenarios beyond image classification, such as object detection and segmentation. Moreover, enhancing the computational efficiency of our method will be critical to its deployment in real-time systems. Ultimately, the goal is to develop a comprehensive framework that addresses a wide range of adversarial challenges, ensuring the safety and reliability of AI-driven technologies.

% TODO add quantit results
 % Experiments on 121 diverse 3D objects emphasize the effectiveness of our approach. In comparison to other techniques, the untargeted Universal attacks successfully identified single noise perturbations with higher destruction rates across multiple viewing angles, particularly at low $\epsilon$ levels. Results showed that between \(\epsilon\) = 0.5 and \(\epsilon\) = 5.0, the untargeted Universal attack displayed the lowest soft-max predictions. In contrast, targeted Universal attacks effectively exhibited its potential for practical adversarial manipulation over a wide range of epsilon values. Targeted Universal attacks exhibited the lowest average soft-max values for test images at .41 all while requiring the least amount of computational effort. This performance emphasizes the improvements over standard single-view attacks, which may struggle with viewpoint variations and deformations while also requiring a noise for each unique viewpoint. Our experiments also substantiates previous research done on the Universal method, validating that the algorithm remains successful on more diverse data.

% \bibliographystyle{dinat}
\bibliography{thbib.bib}

%\if False
%%%%%%%%%%%%%%%%%%%%%%%%%%%%%%%%%%%%%%%%%%%%%%%%%%%%%%%%%%%%

% \newpage
% \newpage

% \appendix

% \section{Appendix / supplemental material}

% \begin{figure*}[h]
%     \centering
%     \includegraphics[width=5.0in,height=2.4in]{images/Drawing.png}
%     \caption{The figure provides a visual representation of how common adversarial attacks fail to incorporate multiple viewpoints.}
%     \label{adv_problem1}
% \end{figure*}

% \begin{figure*}h]
%     \centering
%     \includegraphics[width=5.0in,height=2.4in]{images/Drawing-2.png}
%     \caption{The figure provides a visual representation of how the View-Invariant Adversarial Perturbations can incorporate multiple viewpoints.}
%     \label{adv_problem2}
% \end{figure*}

% Optionally include supplemental material (complete proofs, additional experiments and plots) in appendix.
% All such materials \textbf{SHOULD be included in the main submission.}

%%%%%%%%%%%%%%%%%%%%%%%%%%%%%%%%%%%%%%%%%%%%%%%%%%%%%%%%%%%%
%\fi 

\end{document}